\documentclass[accepted]{uai2023} 

\usepackage{comment}
\usepackage[american]{babel}

\usepackage{natbib} 
\usepackage{mathtools} 
\usepackage{booktabs} 
\usepackage{tikz} 

\usepackage{balance}       
\usepackage[T1]{fontenc}   
\usepackage{txfonts}
\usepackage{mathptmx}
\usepackage{textcomp}
\usepackage{cuted}
\usepackage{tabularx}
\usepackage{bm}
\usepackage[ruled]{algorithm2e}

\usepackage{microtype}        
\usepackage[all]{hypcap}    
\usepackage{ccicons}          
\usepackage[utf8]{inputenc} 

\usepackage{tikz}
\usetikzlibrary{automata,arrows,calc,positioning}

\def\projname{BeliefPPG}

\DeclareMathOperator{\ReLU}{ReLU}

\DeclareMathOperator*{\argmax}{arg\,max}

\newcommand{\oarr}[1]{\overrightarrow{#1}}

\usepackage{xifthen}

\newcommand{\abs}[1]{\left\lvert#1\right\rvert}
\newcommand{\normal}[3][]{%
    \ifthenelse{\isempty{#1}}
        {\mathcal{N}\left(#2, #3\right)}
        {\mathcal{N}\left(#1|#2, #3\right)}%
}

\newcommand\givenbase[1][]{\:#1\lvert\:}
\let\given\givenbase

\DeclarePairedDelimiterX\Basics[1](){\let\given\sgiven #1}

\newcommand\boldblue[1]{\textcolor{gray}{\textbf{#1}}}

\usepackage[flushleft]{threeparttable}

\title{BeliefPPG: Uncertainty-aware Heart Rate Estimation from PPG signals via Belief Propagation}

\author[1]{Valentin~Bieri$^*$}
\author[1]{Paul~Streli$^*$}
\author[1]{Berken~Utku~Demirel}
\author[1]{Christian~Holz}
\affil[1]{%
    Department of Computer Science\\
    ETH Zürich, Switzerland
}

\begin{document}
\maketitle
\def\thefootnote{*}\footnotetext{These authors contributed equally to this work.}\def\thefootnote{\arabic{footnote}}

\begin{abstract}

We present a novel learning-based method that achieves state-of-the-art performance on several heart rate estimation benchmarks extracted from photoplethysmography signals (PPG).
We consider the evolution of the heart rate in the context of a discrete-time stochastic process that we represent as a hidden Markov model.
We derive a distribution over possible heart rate values for a given PPG signal window through a trained neural network.
Using belief propagation, we incorporate the statistical distribution of heart rate changes to refine these estimates in a temporal context.
From this, we obtain a quantized probability distribution over the range of possible heart rate values that captures a meaningful and well-calibrated estimate of the inherent predictive uncertainty.
We show the robustness of our method on eight public datasets with three different cross-validation experiments.

\end{abstract}

\section{Introduction}

Photoplethysmography (PPG) is an optical sensing technique that measures the blood volume pulse from the intensity of reflected light at the surface of the skin.
PPG is widely used in clinical settings to monitor a patient's vital parameters, including heart rate (HR) and pulse oximetry~\citep{PPG-Applications}.
Because PPG sensors are unobtrusive, they are also a common part of wearables such as smartwatches or fitness trackers to monitor cardiac activity and health in ambulatory settings.
The use in ambulatory settings holds much potential for the early diagnosis and prevention of cardiovascular diseases in the general population. 

\begin{figure}[t]
    \centering
    \includegraphics[width=1.0\columnwidth]{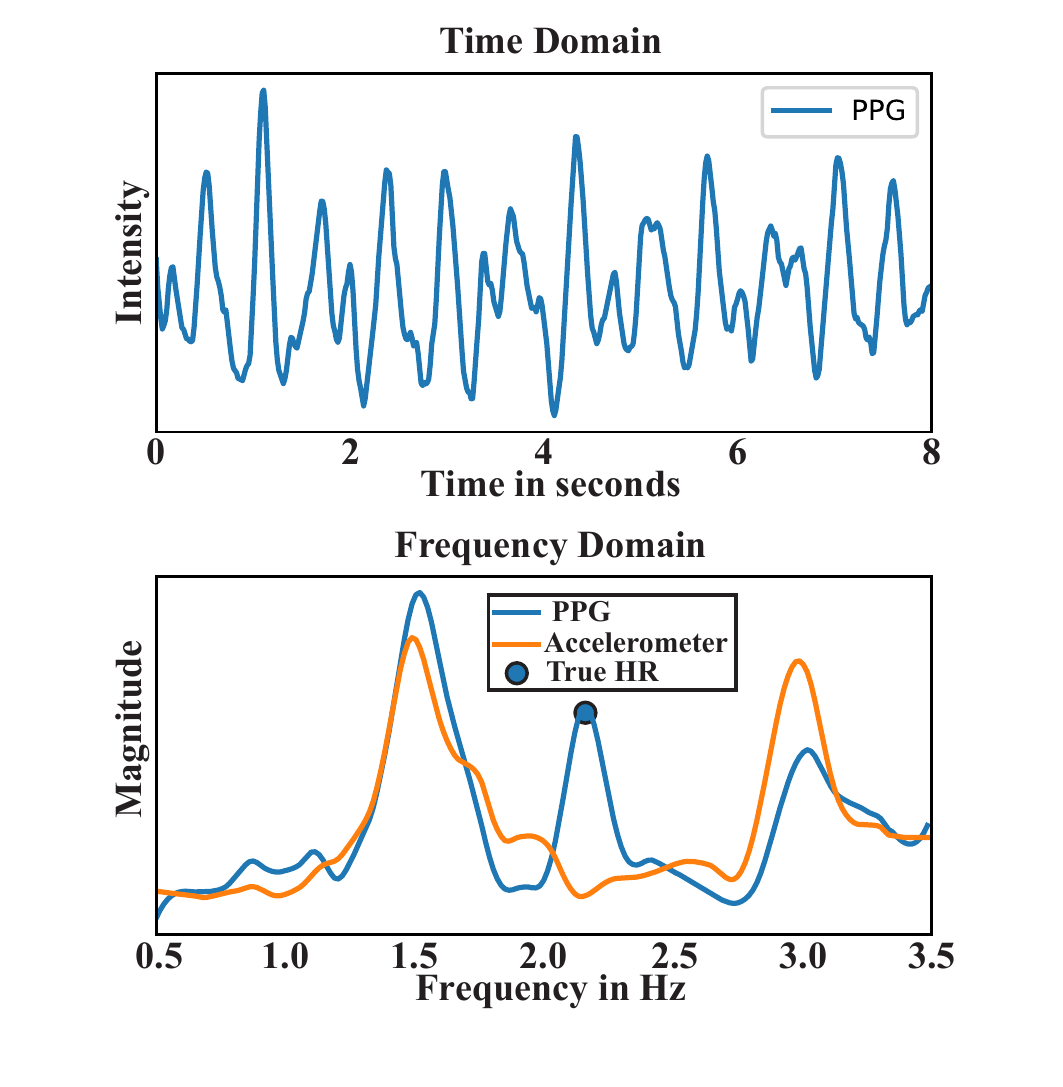}
    \caption{Example of a PPG signal captured with green LED (wavelength:
515\,nm) from the IEEE test dataset~\citep{TROIKA}. Due to motion artifacts, the PPG signal contains several dominant peaks within the range of possible heart rate frequencies. }
    \label{fig:bimodal}
\end{figure}

However, in practice, PPG signals are typically affected by artifacts, especially outside controlled clinical conditions and, thus, where they could prove most useful.
Due to the optical sensing principle, changes in ambient light affect resolved intensities as do motion-induced artifacts from activities such as walking, running, or gesturing.
These motions result in spurious peaks in the signal, causing additional candidates in the frequency spectrum of possible HR values, thus making the extraction of reliable measurements difficult (see \autoref{fig:bimodal}).

Therefore, extracting HR from PPG signals that may be noise-afflicted is an important and challenging problem.
While methods in classical signal processing have long dominated the field of estimating HR from PPG signals, recent research has shown that these algorithms tend to perform poorly on unseen out-of-distribution (OOD) data~\citep{DeepPPG}.
Various deep-learning-based methods have since been proposed~\citep{CorNET, BAMI, DeepHeart}.
Importantly, these methods lack the critical capability to indicate when they fail to produce a reliable result, which is essential in the healthcare sector, where erroneous readings may have severe consequences.

In this paper, we propose a novel learning-based method that simultaneously estimates HR from PPG signals alongside the uncertainty of its estimates.
The key idea of our method is to consider HR prediction within the context of a time-discrete stochastic process.
We model the HR sequence through a probabilistic graphical model that incorporates the statistical transition probabilities between HR classes, such that we can learn a probabilistic relation between instantaneous HR values and observed PPG signal windows through a neural network.
Through message passing, our method takes the probability over all possible past HR trajectories into consideration, allowing us to predict a meaningful distribution over the range of discrete HR classes that captures the predictive uncertainty for an HR estimate at a given time step.
From the mean of the distribution for each time step, we can obtain a single HR estimate. 

Through a series of experiments on eight public datasets with three different cross-validation schemes, we empirically demonstrate the robustness of our proposed method.
For Leave-one-Subject-out (LoSo), we achieve a mean absolute error (MAE) of 3.57 beats per minute (BPM) on PPG-DaLiA~\citep{DeepPPG}, the largest available PPG dataset captured in natural environments.
This outperforms related approaches with state-of-the-art (SOTA) performance by 18\%, even beating them when our method was only trained on samples of other datasets.
We further show that our predicted output distribution is well-calibrated and demonstrate its utility in an experiment where we gradually remove samples with higher predictive uncertainty to further reduce the average prediction error. 

\subsection{Contributions}

We make the following specific contributions in this paper:

\begin{itemize}
    \item a method that estimates heart rate as a hidden stochastic process from a sequence of observed PPG signals, refining the output distribution of a trained neural network through message passing,
    
    \item a novel neural network architecture that operates on both the time and the time-frequency domain for PPG-based HR prediction with fewer trainable parameters than previous methods, and
    
    \item a series of experiments that evaluate the calibration of the predicted output distributions and demonstrate SOTA accuracy of our HR estimates, including on out-of-distribution data\footnote{Our implementation is publicly available at https://github.com/eth-siplab/BeliefPPG.}.  
\end{itemize}

While we demonstrate our method on the problem of PPG-based HR estimation, we believe it could show similarly encouraging results on related temporal sequence estimation problems within other domains.

\section{Related Work}

The work in this paper is related to deep learning-based uncertainty estimation and PPG-based HR estimation.

\label{sec:related_work}

\subsection{Deep learning-based Uncertainty estimation}

The total uncertainty of predictive models can be divided into aleatoric and epistemic uncertainty~\citep{ulmer}.
The aleatoric uncertainty captures the uncertainty inherent in the data and the epistemic uncertainty is an estimate of the uncertainty due to constraints on the model or the training process~\citep{Depeweg2017DecompositionOU}.   
For a well-calibrated model, the uncertainty estimates and the model error are perfectly correlated~\citep{Band2022BenchmarkingBD}.

Several approaches have been proposed to estimate the uncertainty of a deep learning model.
These include fitting a parametric probability distribution over network activations and weights~\citep{model_activation}, sampling using dropout~\citep{old_predictive}, and Bayesian neural networks~\citep{mackay1992practical, neal2012bayesian}. 
Most of the previous work only evaluated their predictive performance and the quality of their uncertainty estimates on curated datasets such as CIFAR-10 and FashionMNIST~\citep{practical_DL,Can_you}.
In the domain of personalized health, where great inter-subject variability is common, robust uncertainty estimates are crucial.

\subsection{PPG-based heart rate estimation}

Increasing interest into PPG-based HR prediction has led to the release of large publicly available datasets~\citep{WESAD, DeepPPG,TROIKA}.
Previous research on PPG-based HR estimation has mainly focused on classical signal-processing methods such as finite state machines (FSM)~\citep{FSM_2} or adaptive filters~\citep{FSM_adaptive,WFPV} that aim to extract the underlying pulse wave component from the PPG signal before estimating HR.
\cite{WFPV} further employed Viterbi decoding to extract the most likely HR path from an extracted histogram, but acknowledges that this approach is constrained to offline post-processing.
\cite{CurToSS} proposed to track heart rate values in the frequency domain using rule-based decision-making algorithms.
However, these methods rely on manually-tuned hyperparameters that do not generalize across datasets where participants' activities differ.

To overcome this, deep learning-based approaches have been introduced for PPG-based HR estimation~\citep{DeepPPG}.
These typically either operate on the time-\citep{CorNET, BinaryCorNET, DeepPulse, PPGNet} or time-frequency \citep{DeepPPG, BAMI} representation of the PPG signal.
Proposed architectures include convolutional, linear and recurrent layers, and have been optimized using Network Architecture Search~ \citep{ActPPG, TEMPONet, NAS-PPG}.
While most of these approaches estimate the most-likely heart rate in a regression setting, \cite{BAMI} reformulates the PPG-based HR estimation as a classification over 222 HR bins.
\cite{DeepPulse} propose a Bayesian neural network combining Bayesian inference with Monte Carlo dropout to obtain estimates of the aleatoric and epistemic uncertainty.
\cite{DeepHeart} and \cite{CardioGAN} use generative machine learning to improve signal quality before applying a classical method.

Our novel deep learning method estimates the HR through a discrete set of classes over the range of possible heart rate values to better represent multi-modal distributions that capture the uncertainty across the frequency spectrum (see~\autoref{fig:bimodal}).
In addition, our method considers its prediction within the temporal context of the HR trajectory using belief propagation to decrease the effect of motion artifacts and OOD samples in an online fashion.

\section{Method}

In this section, we describe our method \projname{} for the reliable estimation of heart rate from PPG signals as input.
We represent the evolution of the heart rate in the form of a probabilistic graphical model.
Using probabilistic inference, our method predicts a representative distribution that captures the inherent predictive uncertainty over the range of possible heart rate values over time.

\subsection{Problem formulation}

Our method estimates the instantaneous heart rate $y_{t}~\in~\mathbb{R}$ at time $t$ based on a window of input signals $\bm{X_{t}}=\left\{\bm{x}_{t-({i-1})T_{x}}\right\}_{i=1}^{L_{x}}$ where $L_{x}\in \mathbb{N}$ is the length and $T_{x}\in\mathbb{R^{+}}$ is the sampling interval of the signal window.
A sample of the signal window $\bm{x}_t\in \mathbb{R}^{N_x}$ generally includes the values of the channels of the PPG sensor as well as the acceleration captured by an accelerometer that is integrated into most sensing hardware for the discrimination of motion artifacts obscuring the PPG signal.

\begin{figure}[b]
    \centering
        \begin{tikzpicture}[node distance={15mm}, 
        main/.style = {thin, draw, minimum size=0.5cm},
        invisible/.style = {minimum size=0.5cm},
        fulledge/.style = {fill, minimum size=0.2cm, node distance={10mm}}
        ] 
            \node[invisible] (11) {\ldots}; 
            \node[main] (12) [right of=11] {$=$};
            \node[main, label=above:$f_{T}$] (13) [right of=12] {};
            \node[main] (14) [right of=13] {$=$};
            \node[invisible] (15) [right of=14] {\ldots};

            \node[main, label=right:$f_{N}$] (22) [below of=12] {};
            \node[main, label=right:$f_{N}$] (24) [below of=14] {};

            \node[fulledge, label=below:$\bm{X_{t-1}}$] (21) [left of=22] {};
            \node[fulledge, label=below:$\bm{X_{t}}$] (23) [left of=24] {};

            \draw[->] (11) -- node[above] {$Y''_{t-1}$} (12) ;
            \draw[->] (12) -- node[above] {$Y_{t-1}$} (13);
            \draw[->] (13) -- node[above] {$Y''_{t}$} (14);
            \draw[->] (14) -- node[above] {$Y_{t}$} (15);
            \draw[<-] (12) -- node[right] {$Y'_{t-1}$}(22);
            \draw[<-] (14) -- node[right] {$Y'_{t}$}(24);
            \draw[-] (21) -- (22);
            \draw[-] (23) -- (24);
        \end{tikzpicture} 
    \caption{Forney factor graph representing the probabilistic graphical model for the evolution of the heart rate over time. The factor graph includes the factor vertices $f_{T}$ and $f_{N}$.
$f_{T}$ captures the transition probabilities between the heart rate states $Y_{t}$ across time.
$f_{N}$ represents the relationship between an observed PPG and acceleration signal window $\bm{X}_{t}$ and the heart rate state $Y_{t}$ at time $t$.}
    \label{fig:factorgraph}
\end{figure}
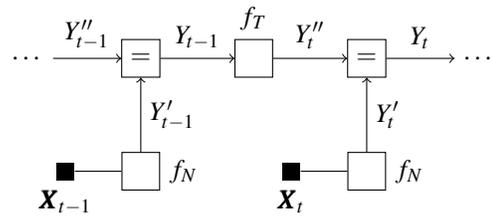

We further consider $y_{t}$ in the context of a discrete-time stochastic process of PPG-based heart rate estimates sampled at an interval $T_y$, where each $y_{t}$ corresponds to an observed signal window $\bm{X_{t}}$ extracted from a recorded trace of PPG and acceleration signals.
We describe this sequence in the form of a Forney factor graph (see~\autoref{fig:factorgraph}), which reveals the structure of a hidden Markov model (HMM) assuming that the hidden heart rate state $Y_{t}$ only depends on its previous state $Y_{t-1}$ and the current observation $\bm{X_{t}}$.
The factor graph comprises the factor vertices $f_{T}$ and $f_{N}$.
$f_{T}$ captures the transition probability between the heart rate state $Y_{t}$ and its previous heart rate state $Y_{t-1}$.
$f_{N}$ models the statistical relation between an observed signal window $\bm{X}_{t}$ and the heart rate state $Y_{t}$ at time $t$, and thus, resembles the emission probability in a standard hidden Markov model.

\subsection{Heart Rate Estimation Network}
We model the factorial $f_{N}\left(Y_{t}, \bm{X_{t}} \right)$ capturing the statistical relation between the heart rate state $Y_{t}$ and the signal window $\bm{X_{t}}$ at time $t$ through a neural network trained with supervision to estimate the conditional probability $p\left(Y_{t} \given \bm{X_{t}}\right)$.
We represent $p_{Y_{t}}\left( \hat{y}_{t} \given \bm{X_{t}}\right)$, $\hat{y}_{t} \in \mathit{C}$, as a discrete distribution over the set of heart rate values $\mathit{C}$ with $c$ classes linearly spaced in the range $[y_{\min},y_{\max})$ to enable the prediction of arbitrarily multimodal distributions~\citep{van2016pixel, BAMI}.
$y_{\min}$ and $y_{\max}$ correspond to physiologically sensible limits of 30\,BPM and 210\,BPM respectively \citep{HR-RANGE}.

\subsection{Heart Rate Transition Function}
The HR transition function is inspired by related medical literature that investigated the distribution of beat-to-beat differences for heart rate variability~\citep{dunnwald2019body, penna1995long, kiyono2004critical}.

We model the heart rate transition factor $f_{T}$ as a conditional probability distribution using the normal distribution fitted to the logarithmic change $\log{\frac{y_{t}}{y_{t-1}}}$ across one timestep (see supplementary material),
\begin{align}\label{eq:f_T}
 p\left(y_{t} \given y_{t-1}\right) = \frac{1}{\sigma_{T}\sqrt{2\pi}}\text{exp}\left({-\frac{1}{2}\left(\frac{\log{\frac{y_{t}}{y_{t-1}}}-\mu_{T}}{\sigma_{T}}\right)^2}\right).
\end{align}

We discretize this conditional probability distribution to the heart rate transition function $f_{T}\left(\hat{y}_{t}, \hat{y}_{t-1}\right)$ by integrating $p\left(y_{t} \given y_{t-1}\right)$ according to the intervals of the heart rate states $\mathit{C}$ from $\hat{y}_{t}^{-}/\hat{y}_{t-1}^{+}$ to $\hat{y}_{t}^{+}/\hat{y}_{t-1}^{-}$, where the superscript $+$ indicates the upper bound and $-$ the lower bound of the interval represented by the respective heart rate class.

\subsection{Probabilistic heart rate inference}

We present two methods for heart rate inference that rely on message passing on the factor graph in~\autoref{fig:factorgraph}.

\subsubsection{HR distribution via belief propagation}
We infer the current heart rate state distribution using forward sum-product message passing (belief propagation), where $\oarr{\mu}$ represents the forward message in the positive time dimension,

\begin{align}
    \oarr{\mu}_{Y_{t}'} \left( \hat{y}_{t} \right) & = f_{N}\left(\hat{y}_{t}, \bm{X_{t}} \right), \\
    \oarr{\mu}_{Y_{t}''} \left( \hat{y}_{t} \right) & = \sum_{\hat{y}_{t-1} \in \mathit{C}} f_{T}\left(\hat{y}_{t}, \hat{y}_{t-1}  \right) \oarr{\mu}_{Y_{t-1}} \left(\hat{y}_{t-1}\right),\\
    \oarr{\mu}_{Y_{t}} \left( \hat{y}_{t}\right) & = \frac{1}{Z_{t}} \oarr{\mu}_{Y_{t}'} \left( \hat{y}_{t} \right)\oarr{\mu}_{Y_{t}''} \left( \hat{y}_{t} \right),\\
    Z_{t} & =  \sum_{\hat{y}_{i}\in\mathit{C}} \oarr{\mu}_{Y_{t}'} \left( \hat{y}_{i} \right)\oarr{\mu}_{Y_{t}''} \left( \hat{y}_{i} \right),  \\
    p_{Y_{t}}\left(\hat{y}_{t} \given \bm{X}_{1},\ldots,\bm{X}_{t} \right) & = \oarr{\mu}_{Y_{t}} \left( \hat{y}_{t}\right).
\end{align}

$p_{Y_{t}}\left(\hat{y}_{t} \given \bm{X}_{1},\ldots,\bm{X}_{t} \right)$ is a probability mass function of the discrete heart rate state $Y_{t}$ over the set of heart rate values $\mathit{C}$.
Note that the forward pass for the computation of $p_{Y_{t}}\left(\hat{y}_{t} \given \bm{X}_{1},\ldots,\bm{X}_{t} \right)$ can be computed in an online manner and efficiently implemented in standard deep-learning frameworks using matrix multiplications.
The distribution over the heart rate classes $p_{Y_{t}}\left(\hat{y}_{t} \given \bm{X}_{1},\ldots,\bm{X}_{t} \right)$ captures the inherent predictive uncertainty of our method.
We can refine the granularity of the discrete heart rate class spacing through linear interpolation and normalization to approximate a more continuous distribution over the range of possible heart rate values.

It can be further reduced to a single heart rate value for time step $t$ through the computation of the mean heart rate, 

\begin{align}
\Tilde{y}_{t} = \mathbb{E}_{\hat{y}_{t}\sim p_{Y_{t}}\left(\hat{y}_{t} \given \bm{X}_{1},\ldots,\bm{X}_{t} \right)}\left[\hat{y}_{t}\right] = \sum_{\hat{y}_{t}\in \mathit{C}} \hat{y}_{t} p_{Y_{t}}\left(\hat{y}_{t} \given \bm{X}_{1},\ldots,\bm{X}_{t} \right).
\end{align}

We compute an estimate of the predictive uncertainty using Shannon's entropy over the distribution 
\begin{align}
\mathcal{H}\left( Y_{t}\right)=-\sum_{\hat{y}_{t}\in \mathit{C}}p_{Y_{t}}\left(\hat{y}_{t} \given \bm{X}_{1},\ldots,\bm{X}_{t} \right)\log{p_{Y_{t}}\left(\hat{y}_{t} \given \bm{X}_{1},\ldots,\bm{X}_{t} \right)}.
\end{align}
We also experiment with the standard deviation over $p_{Y_{t}}\left(\hat{y}_{t} \given \bm{X}_{1},\ldots,\bm{X}_{t} \right)$ to measure how much the distribution is spread out over the heart rate classes.

\subsubsection{HR sequence via max-product message passing}
Alternatively, we can compute an a posteriori estimate of the most likely heart rate sequence over the hidden state variables $\left(\hat{Y}_{1},\ldots,\hat{Y}_{t} \right)$ using max-product message passing that considers the temporal context of the whole sequence, 

\begin{align}
\left(\hat{Y}_{1},\ldots,\hat{Y}_{t} \right) \widehat{=} \argmax_{Y_{1},\ldots,Y_{t}} p\left(Y_{1},\ldots,Y_{t} \given \bm{X}_{1},\ldots,\bm{X}_{t} \right),
\end{align}

where in the case of a unique maximizer

\begin{align}
\hat{y}_{k} = \argmax_{Y_{k}} \max_{Y_{1},\ldots,Y_{t} \text{ except } Y_{k} } p\left(Y_{1},\ldots,Y_{t} \given \bm{X}_{1},\ldots,\bm{X}_{t} \right).
\end{align}

Following the Viterbi algorithm \citep{Viterbi}, we keep track of the product maximizing previous state $\hat{y}_{t-1} \in \mathit{C}$ for each heart rate class $\hat{y}_{t} \in \mathit{C}$ in the forward recursion with the forward message $\oarr{\mu}$ over the heart rate state variables $\left(Y_{1},\ldots,Y_{t}\right)$,

\begin{align}
    \oarr{\mu}_{Y_{t}'} \left( \hat{y}_{t} \right) & = f_{N}\left(\hat{y}_{t}, \bm{X_{t}} \right), \\
    \oarr{\mu}_{Y_{t}''} \left( \hat{y}_{t} \right) & = \max_{\hat{y}_{t-1}\in \mathit{C}} f_{T}\left(\hat{y}_{t}, \hat{y}_{t-1}  \right) \oarr{\mu}_{Y_{t-1}} \left(\hat{y}_{t-1}\right),\\
    \oarr{\mu}_{Y_{t}} \left( \hat{y}_{t}\right) & = \oarr{\mu}_{Y_{t}'} \left( \hat{y}_{t} \right)\oarr{\mu}_{Y_{t}''} \left( \hat{y}_{t} \right).
\end{align}

We then recursively reconstruct and update the maximizing path over the state sequence in a backward pass starting from the maximum heart rate class in the last time step $t$,
\begin{align}
    \hat{y}_{t} & = \argmax_{Y_{t}} \oarr{\mu}_{Y_{t}} \left( \hat{y}_{t}\right), \\
    \hat{y}_{k-1} & =  \argmax_{\hat{y}_{k-1}\in \mathit{C}} f_{T}\left(\hat{y}_{k}, \hat{y}_{k-1}  \right) \oarr{\mu}_{Y_{k-1}} \left(\hat{y}_{k-1}\right).
\end{align}

Due to the backward pass, max-product message passing is more suitable for offline processing. 

\section{Implementation}

We now describe the implementation of our method, including the representation of our input, the HR estimation network, and the HR transition function in detail.

\subsection{Input Representation}
Our network operates on the inputs in their time as well as frequency representation.
The Fourier spectrum of a PPG signal is expected to contain a dominant peak at the heart rate frequency, facilitating a straightforward heart rate estimation for artifact-free signals.
While components from motion artifacts in the PPG signals with frequencies further away from the instantaneous heart rate can be typically eliminated through a comparison with the Fourier spectra of the acceleration channels, leakage and low-resolution effects \citep{stoica2005spectral} in the fast Fourier transform (FFT) can significantly obscure this filtering for motion artifacts with frequencies that are very close to the heartbeat frequency \citep{TROIKA, JOSS}.
The latter are frequently present for wrist-worn sensors during strong movements and repetitive arm motions.
In this case, the patterns in the time-domain representation of the signal might provide additional discerning information, which intuitively motivates the inherent duality of our inputs.

For the time domain representation, we z-score normalize the PPG channels of the window $\bm{X_{t}}$ after band-pass filtering them with corner frequencies 0.1 and 18\,Hz using a Butterworth bandpass filter. Then we average across channels and resample the result to 64\,Hz, obtaining $\bm{\tilde{X}_{t}}\in\mathbb{R}^{L_{x}\times 1}$.

For the Fourier representation, we first subdivide $\bm{X_{t}}$ into another $W_{s}$ overlapping 8-second windows with 2 seconds shift,  z-score normalize and downsample them before computing the FFT with 535 points. To obtain a spectrogram for each window of $\bm{X_{t}}$, we extract the magnitudes of the $N_{s}$ frequency components within the PPG pulse wave frequency range between 0.5\,Hz to 3.5\,Hz. 

We repeat this process for each channel of $\bm{X_{t}}$ separately and then average across available PPG- 
 and acceleration channels to obtain $\bm{\hat{X}_{t}}\in\mathbb{R}^{W_{s}\times N_{s}\times 2}$, where the last dimension consists of the information from the PPG sensor and the accelerometer.

\subsection{Heart Rate Transition Function}
We fit $\mu_{T}$ and $\sigma_{T}$ (see~Eq.~\eqref{eq:f_T}) over the distribution of relative heart rate changes observed in the training set and then perform the discretization depending on the number of heart rate classes $c$. The supplementary material includes an example plot for the heart rate transition distribution extracted from real-world data.

\subsection{Heart Rate Network Architecture}
The architecture of our network consists of two input branches.
One branch operates on the spectrogram $\bm{\hat{X}_{t}}$, while the other receives $\bm{\tilde{X}_{t}}$ as input.

Two consecutive 2D convolutional layers that operate across the time and frequency axes simultaneously with a kernel of size 3x3 extract an embedding for each PPG-acceleration magnitude pair in $\bm{\hat{X}_{t}}$.
A dot-product attention layer then computes the correlation between the linear embeddings computed across the frequency dimension to obtain an attention-refined aggregate, which considers the context across all frequency components along one time step of the spectrogram.
A second dot-product attention layer extracts equivalent features from the spectrogram across the time dimension for each spectrogram frequency.
We add these embeddings to a linear embedding directly computed on the output of the convolutional layers.
This is followed by an average pooling layer that computes the mean of all aggregate vectors along the time dimension. 
A 1D Attention U-Net \citep{UNET-attn} transforms this into a logit vector with $c$ classes.
At the bottleneck of the U-Net, we fuse the latent vector $\bm{h}\in \mathbb{R}^{L_{h}}$ of size ${L_{h}}\in{\mathbb{N}}$ with the output of the time-domain branch.
The time-domain branch consists of a downsized version of CorNET \citep{CorNET}, a two-layer 1D convolutional neural network with max-pooling layers, and a 2-layer LSTM head.
The branch computes a weighting vector $\bm{v}\in \mathbb{R}^{L_{h}}$ and a feature vector $\bm{s}\in \mathbb{R}^{L_{h}}$.

We add $\bm{s}$ and $\bm{v}$ in the form of a residual through an adapted form of additive attention, obtaining the input $\bm{\hat{h}}\in \mathbb{R}^{L_{h}}$ for the upsampling branch of the U-Net as

\begin{equation}
   \bm{\hat{h}} = \bm{h} + \tanh \left( \bm{W_{v}} \left[ \bm{h} ; \bm{v} \right] + \bm{b_{v}} \right) \odot \ReLU \left( \bm{W_{s}} \left[ \bm{h} ; \bm{s} \right] + \bm{b_{s}} \right).
\end{equation}

$\bm{W_{v}}, \bm{W_{s}} \in \mathbb{R}^{L_{h} \times 2L_{h}}$ and $\bm{b_{v}}, \bm{b_{s}} \in \mathbb{R}^{L_{h}}$ are trainable weight matrices and bias terms respectively. $\ReLU$ is the ReLU activation function,
$[\cdot ;\cdot]$ is a concatenation operator and $\odot$ is the element-wise Hadamard product. 

After upsampling, the softmax activation function converts the logit vector to a probability distribution estimate across the $c$ respective classes.
We included a more detailed description of the network architecture in the supplementary material. 

\subsection{Loss function}

We train the network using categorical cross-entropy loss.
Since we reformulate the regression of the real-numbered output $y_{t}$ as a classification problem, we do not use a binary target vector but represent $y_{t}$ as a discrete probability distribution over the set of heart rate values $\mathit{C}$, $p_{Y_{t}}\left(\hat{y}_t\given \bm{X_{t}} \right)$.

We obtain $p_{Y_{t}}\left(\hat{y}_t\given \bm{X_{t}} \right)$ by 
discretizing the normal distribution $\normal{y_{t}}{\sigma_{y}^{2}}$ 
with expectation $y_{t}$ and variance $\sigma_{y}^{2}$.
Here, $y_{t}$ is the ground-truth heart rate and $\sigma_{y}^{2}$ is a hyperparameter, which has a regularizing effect on the output distribution.
Using $p_{Y_{t}}\left(\hat{y}_t\given \bm{X_{t}} \right)$, we compute the categorical cross-entropy loss $\mathcal{L}$ against the estimated output distribution from our network, $\hat{p}_{Y_{t}}\left(\hat{y}_t\given \bm{X_{t}}\right)$ (see \autoref{alg:loss}).

\begin{algorithm}

  \caption{Computation of loss $\mathcal{L}$}
  \label{alg:loss}
  {
    $f\left(y\right) \gets \mathcal{N}(y_t, \sigma_y^2)$

    $p_{Y_{t}}\left(\hat{y}_t\given\bm{X_{t}}\right) \gets f\left(\hat{y}_t\right)/\sum_{\tilde{y} \in \mathit{C}} f\left(\tilde{y}\right)$

    $\mathcal{L} \gets - \sum_{\hat{y}_t\in \mathit{C}} p_{Y_{t}}\left(\hat{y}_t\given\bm{X_{t}}\right)\log{\left(\hat{p}_{Y_{t}}\left(\hat{y}_t\given \bm{X_{t}} \right)\right)}$
  }
  
\end{algorithm}

We use 64 classes, resulting in a mean quantization error $\mathbb{E}\left[\abs{y_{t}-\mathbb{E}_{\hat{y}_{t}\sim p_{Y_{t}}\left(\hat{y}_t\given \bm{X_{t}}\right)}\left[\hat{y}_{t} \right]}\right]$ of 0.02 and a maximum quantization error of around 0.035. 
We implement the heart rate transition function as a matrix $T\in\mathbb{R}^{c \times c}$ to ease integration in the message-passing algorithm.

\subsection{Training}
We implement our model in TensorFlow. 
The models are trained on an \textit{NVIDIA GeForce RTX 3090} GPU until convergence with an early stopping criterion obtained on a separate validation set.
We use a batch size of 128 and the Adam~\citep{ADAM} optimizer with a learning rate of $2.5 \times 10^{-4}$.
We further augment our training dataset by applying random time stretches with up to 25\% on the input signals and extracted labels as well as  
by adding Gaussian noise $n$ to the inputs, $n \sim \normal{0}{0.25^{2}}$.

\section{Experiments}

We evaluate our approach on eight publicly available datasets using three different cross-validation schemes to test for robustness and generalization capability. 
 Mean absolute errors (MAEs) and standard deviations are presented as the average and standard deviation across sessions.

\subsection{Cross-Validation Schemes}
\paragraph{Leave-one-Session-out (LoSo) cross-validation}
 The subjects are split into $n-2$ training, one validation, and one test subject.
 The split is rotated until all sessions have been used as test set. We perform the experiments on each dataset separately.
\par
\paragraph{Five-Fold cross-validation} Each dataset is randomly split into 20\% test and 80\% training sessions, where each session captures the recording for a unique subject.
One validation subject is randomly chosen from the training set.
We combine the respective splits from all datasets so that every dataset contributes to the experiment for each fold. 

\par
\paragraph{Leave-one-Dataset-out cross-validation} One dataset is taken as test data, and the remainder serves as training data. One validation session is drawn from each training dataset. 
\par
Only predictions on test sets (unseen data) are reported.
The model is optimized on the training set until validation loss does not decrease for more than 50 epochs.

\subsection{Datasets}
Our evaluation considers eight datasets, including IEEE train and test~\citep{TROIKA}, PPG-DaLiA~\citep{DeepPPG}, WESAD~\citep{WESAD}, the two datasets published by the BAMI-Labs~\citep{BAMI, BAMIDS}, BIDMC PPG~\citep{BIDMC} as well as CLAS~\citep{CLAS}.
They jointly provide a diverse data collection comprising different subjects, activities, and recording hardware. 
DaLiA is not only by far the largest but also the only data collection gathered in natural environments.

\subsection{Heart rate estimation accuracy}
\label{sec:results}

\begin{table*}[t]
\centering
\caption{HR estimation accuracy compared to prior work. Values indicate MAE and standard deviation in BPM using LoSo cross-validation if not indicated by footnotes otherwise.}
\label{fig:resulttable}
\resizebox{\textwidth}{!}{
     \begin{threeparttable}
    \begin{tabular}{llcccccccc} \toprule
        ~ & ~ & ~ & ~ & ~ & ~ & ~ & ~ & ~ & ~ \\ 
        \textbf{Approach} & \textbf{\#Param.} & \textbf{IEEE train} & \textbf{IEEE test} &\textbf{PPG-DaLiA} & \textbf{WESAD} & \textbf{BAMI-1} & \textbf{BAMI-2} & \textbf{BIDMC} & \textbf{CLAS} \\ 
         \midrule
        \multicolumn{3}{l}{\textbf{CLASSICAL METHODS} (not cross-validated)} & ~ & ~ & ~ & ~ & ~ & ~ & ~ \\ 
        WFPV\tnote{1} ~\citep{WFPV} & - & \textbf{1.02} & \textbf{1.97} & 10.7 ± 3.8 & 8.5 ± 4.3 & 11.28 & 6.09 & - & - \\ 
        TAPIR\tnote{1} ~\citep{TAPIR} & - & 2.5 ± 1.2 & 5.9  ±  3.5 & \phantom{0}4.6 ± 1.4 & 4.2 ± 1.4 & - & - & - & - \\ 
        CurToSS\tnote{1} ~\citep{CurToSS} & - & 2.2 & 4.5 & \phantom{0}5.0 ± 2.8 & 6.4 ± 1.8 & - & - & - & - \\ 
        ~ & ~ & ~ & ~ & ~ & ~ & ~ & ~ & ~ & ~ \\ 
        \multicolumn{3}{l}{\textbf{DEEP LEARNING METHODS}} & ~ & ~ & ~ & ~ & ~ & ~ & ~ \\ 
        CardioGAN~\citep{CardioGAN} & - & - & - & 8.30 & 8.60  & - & - & \textbf{0.7} & - \\ 
        DeepHeart~\citep{DeepHeart} & 3.3M & 4.76 \textit{(1.98\tnote{6} )} & - & - & - & - & - & - & - \\ 
        BAMI\tnote{2} ~\citep{BAMI} & 3.3M & \textit{0.67}\tnote{4}  & \textit{0.86\tnote{4}} & - & - & \textit{1.39\tnote{4}} & 1.46 & - & - \\ 
        Binary CorNET~\citep{BinaryCorNET} & 257k & 4.67 ± 3.7 & \phantom{0}6.61 ± \phantom{0}5.4 & - & - & - & - & - & - \\ 
        PPGNet~\citep{PPGNet} & - & 3.36 ± 4.1 & 12.48 ± 14.5 & - & - & - & - & - & - \\
        Deep PPG~\citep{DeepPPG} & 8.5M & 4.0\phantom{0} ± 5.4 & 16.51 ± 16.1 & 7.65 ± 4.2 & 7.47 ± 3.3 & - & - & - & - \\ 
        DeepPulse~\citep{DeepPulse} & 730k & \textit{2.76 ± 3.0\tnote{5}} & \textit{5.05 ± 5.5\tnote{5}} & \textit{2.12 ± 3.1\tnote{5}} & - & - & \textit{\phantom{..0}2.38 ± 2.6\tnote{5}} & - & - \\ 
        NAS-PPG\tnote{3} ~\citep{NAS-PPG} & 800k &\textit{0.82}\tnote{4} & \textit{1.03}\tnote{4} & \phantom{0}6.02 ± 10.6 & - & - & - & - & - \\ 
        MH Conv-LSTM~\citep{MultiHeadedConv} & 680k & - & - & 6.28 ± 3.5 & - & - & - & - & - \\ 
        ActPPG\tnote{3} ~\citep{ActPPG} & 900k & \multicolumn{2}{c}{3.27 ± 2.0} & 4.88 (\textit{3.84}\tnote{7} \phantom{.}) & - & - & - & - & - \\ 
        TEMPONet\tnote{3}~\citep{TEMPONet} & 269k & - & - & 4.36 (\textit{3.61}\tnote{7} \phantom{.}) & - & - & - & - & - \\ 
        ~ & ~ & ~ & ~ & ~ & ~ & ~ & ~ & ~ & ~ \\ 
        \multicolumn{3}{l}{\textbf{OURS}} & ~ & ~ & ~ & ~ & ~ & ~ & ~ \\ 
        \multicolumn{3}{l}{\textit{Leave-one-Session-out Cross-Validation}}   & ~ & ~ & ~ & ~ & ~ & ~ & ~ \\
        \projname{} (ours) & \textbf{138k} & 1.75 ± 0.8 & 3.78 ± 2.2 & 3.57 ± 1.4 & 4.28 ± 2.0 & \textbf{2.00 ± 1.0} & 1.48 ± 0.9 & - & - \\ 
        \projname{} / viterbi &  \textbf{138k} & \boldblue{1.47 ± 0.6} & 3.06 ± 1.9 & \textbf{3.18 ± 1.3} & 4.02 ± 1.9 & \boldblue{2.12 ± 1.1} & \boldblue{1.46 ± 0.3} & - & - \\ 
        \multicolumn{3}{l}{\textit{Five-Fold Cross-Validation}} & ~ & ~ & ~ & ~ & ~ & ~ & ~ \\
        \projname{} (ours) & \textbf{138k} & 1.82 ± 0.6 & 4.20 ± 2.3 & 3.74 ± 1.8 & 3.81 ± 1.9 & 2.17 ± 1.1 & \textbf{1.32 ± 0.3} & \boldblue{1.11 ± 1.7} & \boldblue{1.47 ± 1.0} \\ 
        \projname{} / viterbi &  \textbf{138k} & 1.49 ± 0.4 & \boldblue{2.61 ± 1.4} & \boldblue{3.34 ± 1.7} & \textbf{3.63 ± 1.9} & 2.20 ± 1.0 & 1.53 ± 0.2 & 1.40 ± 1.5 & 1.48 ± 0.8 \\ 
        \multicolumn{3}{l}{\textit{Leave-one-Dataset-out Cross-Validation}}  & ~ & ~ & ~ & ~ & ~ & ~ & ~ \\
        \projname{} (ours) & \textbf{138k} & 1.88 ± 0.7 & 4.39 ± 3.1 & 4.26 ± 1.6 & 3.88 ± 1.9 & 3.83 ± 3.8 & 2.90 ± 3.5 & 1.34 ± 1.6 & \textbf{1.41 ± 0.8} \\ 
        \projname{} / viterbi &  \textbf{138k} & 1.48 ± 0.5 & 3.14 ± 2.1 & 3.83 ± 1.6 & \boldblue{3.70 ± 1.8} & 3.67 ± 3.7 & 2.96 ± 3.5 & 1.59 ± 1.5 & 1.67 ± 0.8 \\ 
        \bottomrule
    \end{tabular}
\begin{tablenotes}
            \item[1] Reported results without cross-validation of hand-tuned parameters
            \item[2] Reported results using a single train-test split across datasets
            \item[3] Performed network architecture search with respect to test set            
            \item[4] Training error
            \item[5] Mixed training and testing error 
            \item[6] Applies post-processing technique without cross-validation
            \item[7] Subject-specific fine-tuning
        \end{tablenotes}
    \end{threeparttable}
}

\end{table*}

On all eight datasets and under all three evaluation strategies the method achieves or outperforms state-of-the-art results among cross-validated methods.
A full comparison is shown in Table~\ref{fig:resulttable}. 
Under leave-one-session-out as well as five-fold cross-validation the model achieves low mean absolute errors despite the small dataset sizes in the former and the heterogeneous data in the latter. On PPG-DaLiA the method outperforms the reported MAE of the runner-up method Q-PPG by 0.8 (18\%) BPM under LoSo evaluation.
The small IEEE datasets however remain challenging: Although achieving the lowest error among LoSo-evaluated deep learning methods, hand-tuned algorithms such as WFPV still perform slightly better.
\par 
During out-of-distribution evaluation (Leave-one-Dataset-out) we observe a slight decrease in accuracy on the DaLiA and BAMI datasets (+0.5, +1.7, +1.6 BPM MAE) while the performance on the other datasets remains unchanged.
Notably, our method still outperforms its in-distribution-trained competition on PPG-DaLiA.
\par 
In summary, the model produces highly competitive results in terms of MAE under various data distributions and cross-validation methods. 

\par

\begin{figure}[t]
    \centering
    \includegraphics[width=1.0\columnwidth]{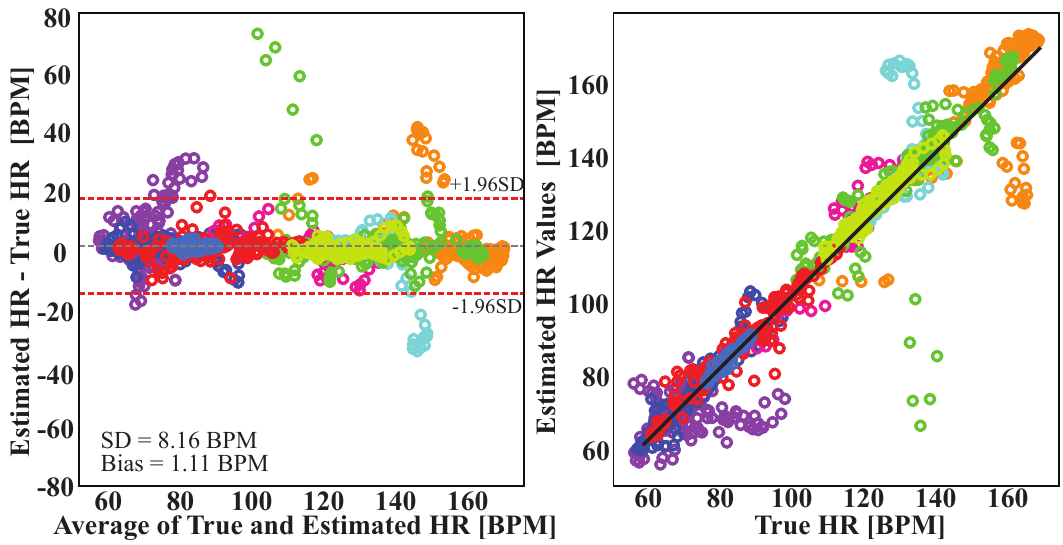}
    \caption{Bland-Altman plot of HR estimates during LoSo cross-validation on IEEE test. The colors represent subjects.}
    \label{fig:bland_alt}
\end{figure}

\begin{figure}[t]
    \centering
    \includegraphics[width=1.0\columnwidth]{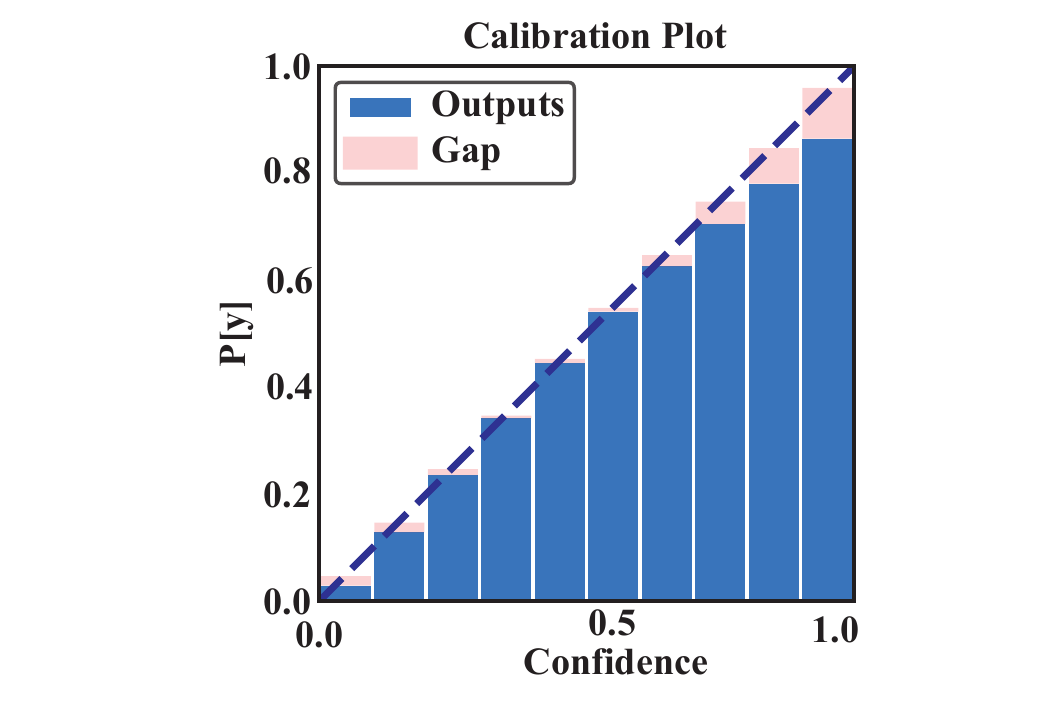}
    \caption{Model calibration visualized using the empirical probability of ground truth HR being inside the confidence region. The model is slightly overconfident for high-confidence levels, but shows good overall calibration.}
    \label{fig:calib_plot}
\end{figure}

\paragraph{Errors by activity}
We added a plot of the relative errors achieved across activities performed in PPG-DaLiA in the supplementary material. We observe the highest errors for the "walking stairs" activity (6.6\% mean absolute percentage error) and the lowest errors for sedentary activities.

\paragraph{Errors by subject}
As shown in Table \ref{fig:resulttable}, the standard deviation of the MAE across subjects is low in comparison with existing methods. Figure~\ref{fig:bland_alt} further plots the error on IEEE test by subject.

\subsection{Evaluating Predictive Uncertainty}

\paragraph{Calibration} 
To evaluate the calibration of our predicted heart rate distributions, we assess whether the empirical probability of the ground truth heart rate falling within a heart rate region is matching the regions aggregated probability mass.
In our experiment, we successively consider the intervals representing the class bins from the heart rate class with the highest assigned probability value to the heart rate class with the lowest probability value.

Results are plotted in a calibration plot in Figure~\ref{fig:calib_plot}, which shows that our method is generally well-calibrated, but slightly overconfident for higher confidence levels.

We also present the negative log-likelihoods (NLL) of our method on the test set as a measure of its predictive performance in~\autoref{fig:log_lik_table}.

\begin{table}[h]
\centering
\caption{NLL on the test set for \projname{} under various Cross-Validation (CV) Schemes.}
\label{fig:log_lik_table}
\resizebox{\columnwidth}{!}{
    \begin{tabular}{lcccccc} \toprule
        ~ & \textbf{IEEE train} & \textbf{IEEE test} & \textbf{DaLiA} & \textbf{WESAD} & \textbf{BAMI-I}& \textbf{BAMI-II} \\ 
        \midrule
        Leave-one-session-out CV &  3.92  & 5.07 & 4.74 & 4.89 & 4.12 & 3.82 \\ 
        Five-fold CV & 4.01 & 4.82  & 4.78 & 4.7 & 4.18 &3.75 \\ 
        Leave-one-dataset-out CV & 4.08 & 4.83 & 5.01 & 4.89 & 4.68 & 4.19\\ 
        \bottomrule
    \end{tabular}
}
\end{table}

\begin{figure}[t]
    \centering
    \includegraphics[width=1.0\columnwidth]{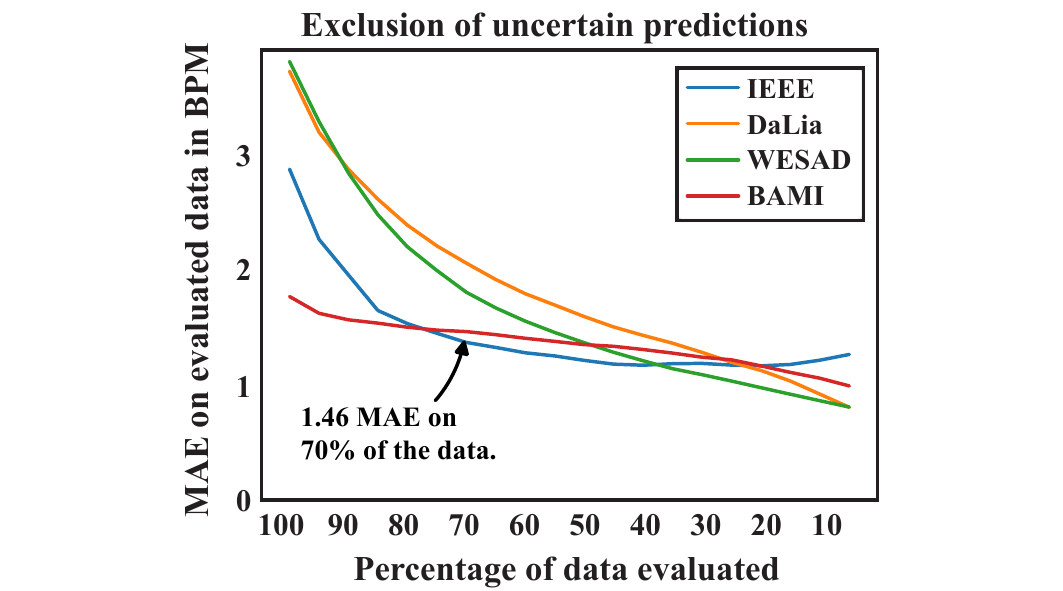}
    \caption{Evolution of MAE for five-fold cross-validation while successively excluding the most uncertain predictions, using entropy as metric.}
    \label{fig:case_study}
\end{figure}

\paragraph{Quantification of uncertainty}
The estimated probability distribution captures the predictive uncertainty of our model.
This can be used to discard samples with high uncertainty.
To that end, we quantify the predictive uncertainty using entropy and successively discard the most uncertain predictions while computing the MAE on the remaining samples. By removing the top-1\% most uncertain predictions (over all datasets and experiments) we reduce MAE by 4\%; by removing the top-5\% we achieve an improvement of 15\%. 
Figure~\ref{fig:case_study} visualizes the effect on different datasets.
Standard deviation and entropy achieve almost equal (within $\sim$1\%) results in these experiments, indicating that they both represent meaningful estimates of the predictive uncertainty.

The calibration was evaluated after linearly upsampling the probability distributions to 1000 intervals.

\subsection{Ablation study}

\begin{table}[ht!]
\centering
\caption{Results (MAE in BPM) of ablation experiments with five-fold CV. The first row corresponds to a version of \projname{} with a higher resolution of both features and labels, resulting in an additional layer of depth in the U-Net component. Subsequently, the table presents incremental ablations of the final model (bold). Note that the last three rows represent models that operate on a shorter feature window of eight seconds, resulting in a single timestep in the spectral representation ($W_s = 1$ instead of $W_s = 7$), thereby rendering the initial dot-product attention layer obsolete. U-Net and CorNET Light refer to the standalone Fourier and time-domain branches of the model, respectively.
}

\label{fig:ablation_table}
\resizebox{\columnwidth}{!}{
    \begin{tabular}{lcccccc} \toprule
        ~ & \textbf{IEEE train} & \textbf{IEEE test} & \textbf{PPG-DaLiA} & \textbf{WESAD} & \textbf{BAMI-I} & \textbf{BAMI-II} \\ 
        \midrule   
        \textit{\projname{} (256 bins)} & 1.66 $\pm$ 0.8 & 4.53 $\pm$ 4.0 & 3.65 $\pm$ 1.4 & \textbf{3.89 $\pm$ 1.9} & \textbf{1.96 $\pm$ 1.0} & \textbf{1.24 $\pm$ 0.4} \\
        \textbf{\projname{} (64 bins)} & 1.67 $\pm$ 0.6 & \textbf{4.47 $\pm$ 2.9} & \textbf{3.59 $\pm$ 1.7} & 3.94 $\pm$ 2.1 & 2.08 $\pm$ 1.0 & 1.34 $\pm$ 0.3 \\
        - w/o Message Passing & \textbf{1.62 $\pm$ 0.8} & 7.61 $\pm$ 5.2 & 4.02 $\pm$ 1.9 & 3.98 $\pm$ 2.0 & 2.42 $\pm$ 1.2 & 1.4 $\pm$ 0.3 \\ 
        - w/o Data Aug. & 1.62 $\pm$ 1.1 & 8.56 $\pm$ 8.9 & 4.02 $\pm$ 1.9 & 4.17 $\pm$ 2.0 & 2.43 $\pm$ 1.4 & 1.5 $\pm$ 0.8 \\ 
        - with 8-second window & 1.69 $\pm$ 1.2 & 13.27 $\pm$ 10.0 & 4.86 $\pm$ 2.3 & 4.6 $\pm$ 2.2 & 3.32 $\pm$ 1.6 & 1.97 $\pm$ 0.9 \\ 
        U-Net (8-sec) & 2.65 $\pm$ 1.9 & 11.16 $\pm$ 8.1 & 7.93 $\pm$ 4.2 & 6.89 $\pm$ 2.7 & 3.21 $\pm$ 1.5 & 1.84 $\pm$ 0.7 \\ 
        CorNET Light (8-sec) & 9.85 $\pm$ 7.7 & 19.32 $\pm$ 9.9 & 5.31 $\pm$ 2.7 & 4.24 $\pm$ 2.0 & 9.86 $\pm$ 6.7 & 3.16 $\pm$ 1.8 \\ 
        \bottomrule
    \end{tabular}
}

\end{table}

We perform an ablation for our proposed method.

\par 
\paragraph{Network Architecture} Performance degrades when successively disassembling the proposed architecture. Table~\ref{fig:ablation_table} shows that the omission of the data augmentation leads to worse results. The additional removal of the attention layer that combines the seven frequency-timesteps increases errors by more than 20\% on PPG-DaLiA. Notably, this reduces the overall considered signal window to 8s. If we further decompose the resulting network into the downsized CorNET and the frequency-domain U-Net, MAEs increase by another 9\% and 63\% respectively.

\paragraph{Effect of message passing} Table \ref{fig:ablation_table} also shows how temporal modeling in the form of message passing improves performance. The effect is particularly strong on datasets with larger errors such as IEEE test and PPG-DaLiA.

\paragraph{Output HR resolution} We validate the chosen tradeoff between complexity and accuracy in representing the output space with 64 intervals (2.81 BPM width) by running experiments against a higher resolution of 256 intervals (0.70 BPM width). The input spectra and the U-Net are scaled by the same factor, resulting in a non-trivial increase in the computational cost. Results indicate little improvement in accuracy.

\paragraph{Regularization with $\sigma_y$} To reduce the quantization error in the label we use a Gaussian distribution with a standard deviation  of $\sigma_y$ = 1.5. Ablations (provided in the supplementary material) indicate that the choice of $\sigma_y$ has very little impact on performance.

\subsection{Offline Decoding}
If online estimates are not required, Viterbi decoding can further decrease error. On PPG-DaLiA, this results in an additional reduction of  11\% MAE compared to online decoding. Corresponding results are displayed in Table \ref{fig:resulttable}.

\par

\section{Limitations and future work}
The HR discretization introduces quantization errors, albeit small, and imposes additional hyperparameters on the model such as the number of bins, label variance $\sigma_y^2$, and the fixed range. Experiments, however, demonstrate low sensitivity towards these parameters. 
Despite not being specific to HR estimation, the framework is limited to low-variate regression problems due to the exponential scale-up of the number of output classes. 

We anticipate that our approach can be readily adapted to address a broad range of related problems involving 1D-time signals exhibiting a similar HMM structure, including various other health-related challenges~\citep{karlen2021capnobase, pino2017bcg, chen2018deepphys,kang2018novel}.
We recognize these as promising directions for future research.
Furthermore, considering the utilization of self-supervised learning techniques~\citep{ghorbani2022self, yang2023simper} with large-scale unlabeled PPG datasets, collected from wearable devices in real-world settings without ground-truth labels, may lead to improved outcomes.

\section{Conclusion}

We have introduced a novel deep-learning method for PPG-based heart rate estimation.
Our method joins the time and the time-frequency domains with a dual, U-Net-based neural network, producing discrete probability distributions over HR intervals.
These distributions are contextualized through belief propagation over time, allowing online inference of expected heart rate.

Through extensive experiments on eight publicly available datasets, we have demonstrated state-of-the-art performance and robustness of our method, surpassing previous methods on PPG-DaLiA by 18\% in accuracy.
Even in our out-of-distribution evaluation, our model outperforms existing methods.
The predictive uncertainty of our method is well-calibrated, in particular in lower certainty levels.
Measuring uncertainty by entropy enables incremental MAE reduction by rejecting uncertain predictions, with 5\% rejections resulting in a 15\% boost on average.
Alternatively, performance can be improved by max-product message passing offline.

We believe that our proposed method has the potential to not only enhance the reliability of cardiac monitoring but also address various related problems involving quasi-periodic 1D-time signals.

\bibliography{bieri_741}
\end{document}


\maketitle

\def\thefootnote{*}\footnotetext{These authors contributed equally to this work.}\def\thefootnote{\arabic{footnote}}

\appendix
\section{Transition function selection}
\label{sec:transition_fn}
\begin{figure}[ht!]
    \centering
    \includegraphics[width=0.7\columnwidth]{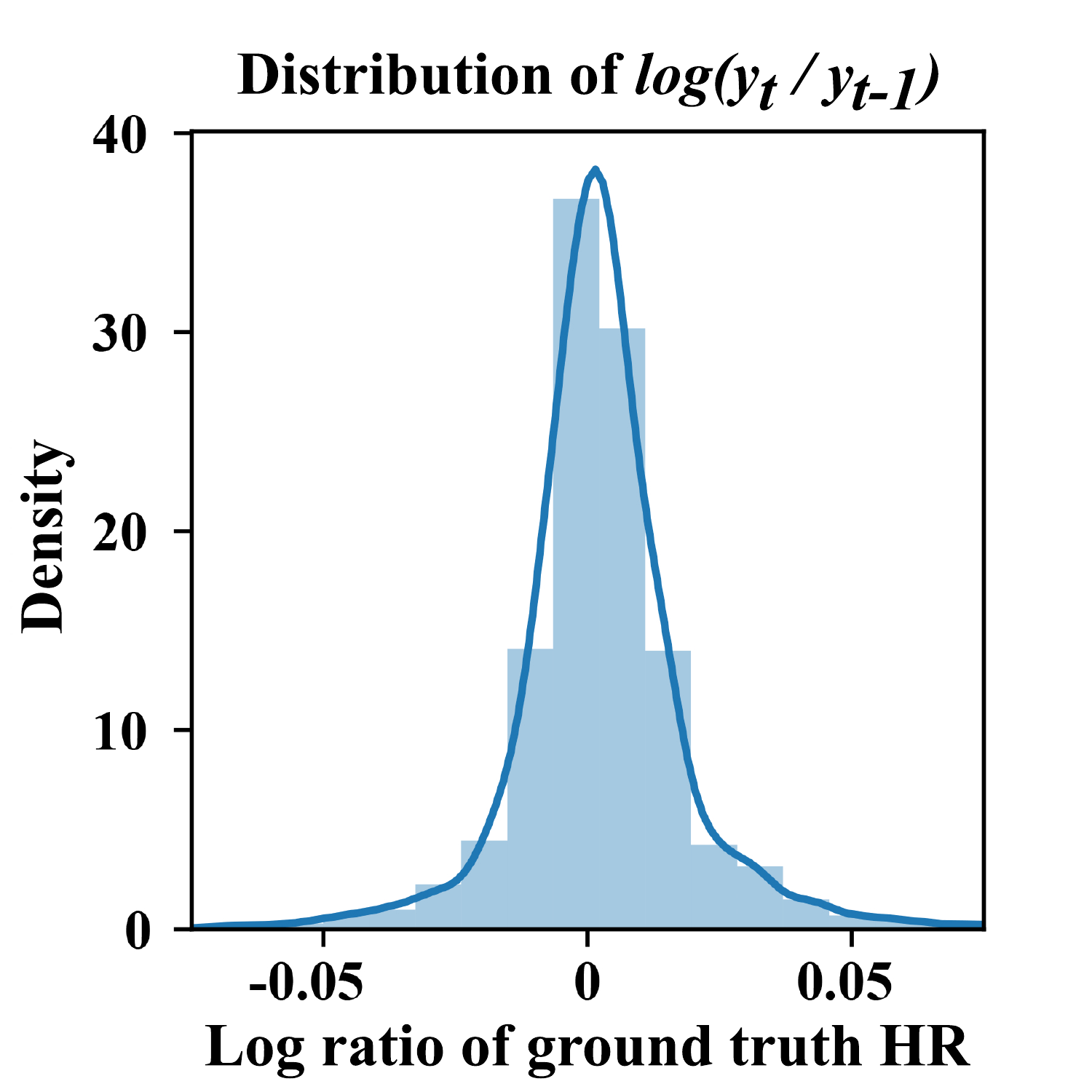}
    \caption{Logarithmic change of HR values in the IEEE dataset.}
    \label{fig:logdiffs}
\end{figure}

While we experimented with alternative distributions for the transition function, including LaPlacian, Gaussian, and Levy distributions of the absolute difference and relative difference in HR between two beat-to-beat intervals, we observed that our transition function using a discretized Gaussian prior fit on the logarithmic change $\log{\frac{y_{t}}{y_{t-1}}}$ led to the best results while offering a reasonable fit to the observed histogram of heart range changes (see~\autoref{fig:logdiffs}).

\section{Selection of $\sigma_{y}$}
\label{sec:suppl_ablation}
We represent the ground-truth HR $y_{t}$ as a discretized normal distribution with density $\normal{y_{t}}{\sigma_{y}^{2}}$. The selection of the hyperparameter $\sigma_y^2$ is based on an ablation study, and the results, as demonstrated in Table~\ref{fig:label_std_table}, indicate minimal sensitivity to variations in $y_t$.

\begin{table}[h!]
\centering
\caption{Ablations of \projname{} with different $\sigma_y$ under LoSo-CV on BAMI-1. We report MAE, its standard deviation across subjects (STD-MAE) and the negative log-likelihood (NLL) on the test set.}
\label{fig:label_std_table}
\resizebox{\columnwidth}{!}{
    \begin{tabular}{lccccccccc} \toprule
        $\sigma_y$ & 0.25 & 0.5 & 0.75 & 1.0 & 1.5 & 2.0 & 2.5 & 3.0 & 4.0  \\ \midrule
        MAE & 2.15 & 2.30 & 2.34 & 2.18 & \textbf{2.00} & 2.45 & 2.29 & 2.45 & 2.35 \\
        STD MAE & 0.90 & 1.30 & 1.70 & 1.00 & \textbf{1.00} & 1.80 & 1.50 & 1.50 & 1.40 \\
        \textbf{NLL} & 4.13 & 4.2 & 4.22 & 4.13 & \textbf{4.12} & 4.28 & 4.23 & 4.3 & 4.33 \\ \bottomrule
    \end{tabular}
}
\end{table}

\section{MAPE across activities}

\begin{figure}[h!]
    \centering
    \includegraphics[width=1.0\columnwidth]{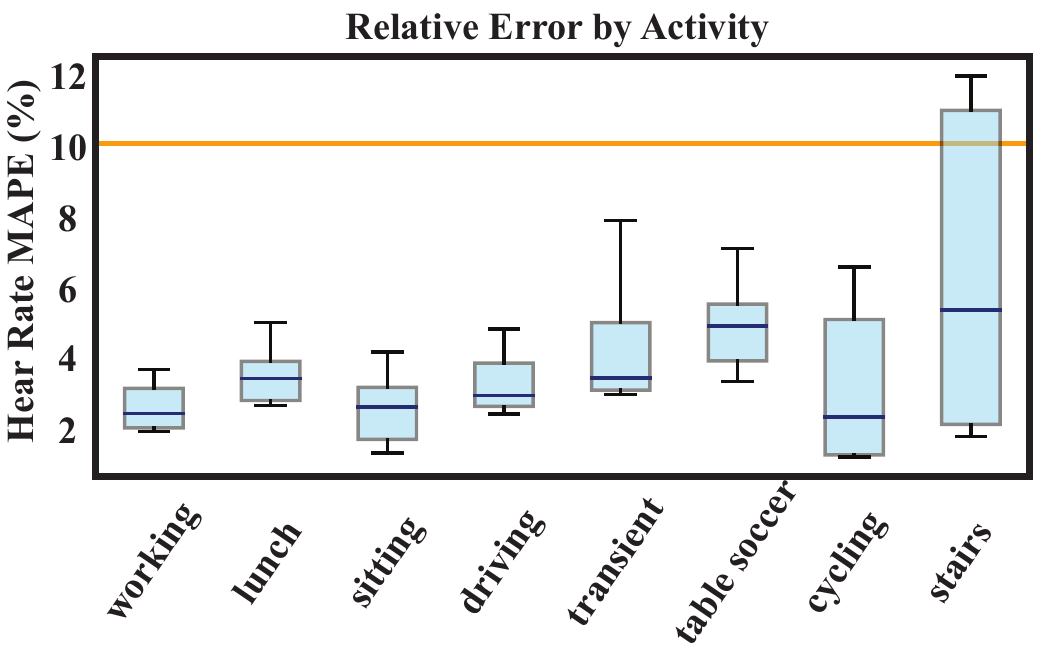}
    \caption{Mean absolute percentage errors (MAPEs) visualized by activity for LoSo on DaLiA. Boxes correspond to interquartile ranges and whiskers to 10th and 90th percentiles over subjects. The blue lines indicate the median, and the orange line marks the acceptable error of 10\%.}
    \label{fig:by_activity}
\end{figure}
Figure \ref{fig:by_activity} presents the Mean Absolute Percentage Errors (MAPE) across activities compared to the AAMI standard \citep{AAMI} for the DaLiA dataset.
The AAMI standard sets the acceptable limits for HR monitoring within ±10\%, which is implemented using the MAPE statistic as defined by the Consumer Technology Association \citep{Consumer-technology-assoc}.

The results demonstrate that the median MAPE for all activities is significantly below this limit. Notably, none of the 15 peak detectors benchmarked by Charlton et al. \citep{CharltonBenchmark} achieved this level of accuracy.

\section{Architecture details}
\label{sec:nn_config}
Table \ref{fig:architecture_table} provides architecture and training details. We implemented our network using TensorFlow 2.8.0.

\begin{table*}[t!]
\centering
\caption{Details about \projname{}'s heart rate network architecture and training configuration}
\label{fig:architecture_table}

\resizebox{\textwidth}{!}{
    \begin{tabular}{llllllllllll}
        \textbf{} & \textbf{} & \textbf{} &   \\ \hline
        \multicolumn{3}{l}{\textbf{Heart rate network architecture}} ~ & ~ & ~ & ~ & ~ & ~& ~ \\ 
        \textbf{} & Up-/Downspl. fac. & Comment & Kernel Size & Padding & Filters / Inner Dim & Activation & Dropout & Output shape & ~ & ~ & ~ \\ 
        
        \midrule
        
        \multicolumn{3}{l}{\textbf{Time-Frequency Branch} \textit{ // input shape: $(W_s,N_s,2) = (7, 64, 2)$}}  ~ & ~ & ~ & ~  \\ 
         {2x Conv2D} & - & - & 3x3 & same & 32 & leaky\_relu & 0.1 & (7,64,32) \\ 
         {embedding 4x} & - & - & - & - & 32 & - & - & 4 x (7,64,32) \\ 
         {2x attention + reduce mean}  & - & reduces $1^{st}$ axis & - & - & - & - & - & (64, 32) \\ 
         {3x downsampling block (1D)} & 4 & stride=poolsize=4 & 3x1 & same & 12, 24, 48 & relu & 0.2 & (1, 48)  \\ 
         {bottleneck attention} & - & - & 2x1 & - & 48 & tanh, relu & 0.2 & (1, 48)\\ 
         {3x upsampling block} & 4 & upspl. size=4 & 3x1 & same & 48, 24, 12 & relu & 0.2 & (64, 12) \\ 
         {Conv1D} & - & - & 1x1 & same & 1 & softmax & - & (64,) \\ 
         {} & ~ & ~ & ~ & ~ & ~ & ~ \\ 
         \multicolumn{3}{l}{\textbf{Time Branch} \textit{ // input shape: $(L_x, 1) = (1280, 1)$}} & ~ & ~ & ~  \\ 
        {Conv1D + bn + MaxPool} & 4 & dilation\_rate=2 & 10 & causal & 16 & leaky\_relu & 0.1 & (320, 16)  \\ 
        {Conv1D + bn + MaxPool} & 4 & dilation\_rate=2 & 10 & causal & 16 & leaky\_relu & 0.1 & (80, 16)  \\ 
        {LSTM} & - & - & - & - & 64 & tanh & 0.1 & (80, 64)\\ 
        {LSTM} & - & - & - & - & 64 & tanh & 0.1 &  (64,) \\ 
        {Dense 2x} & - & - & - & - & 48 & leaky\_relu & - & 2x (1, 48) \\ 
        \textbf{} & ~ & ~ & ~ & ~ & ~ & ~ & ~  \\ 
        \multicolumn{3}{l}{\textbf{Training Parameters}} ~ & ~ & ~ & ~ & ~  \\ 
        {Batch Size} & 128 & ~ & ~ & ~ & ~ & ~ & ~ & ~\\ 
        {Optimizer} & \multicolumn{5}{l}{Adam(lr=0.00025)} ~ & ~ & ~ & ~ \\ 
        {LR Scheduler} & \multicolumn{6}{l}{ReduceLROnPlateau(factor=0.5, min\_lr=1e-10 monitor="loss", patience=3)} ~ & ~ & ~ \\ 
        {Stopping criterion} & \multicolumn{5}{l}{EarlyStopping(patience=40 restore\_best\_weights=True, monitor="val\_loss")} ~ & ~ & ~  \\ 
        {TTA: gaussian noise std} & 0.25 & ~ & ~ & ~ & ~ & ~ & ~ & ~ \\ 
        {TTA: max stretching factor} & 25\% & ~ & ~ & ~ & ~ & ~ & ~ \\  
        {$\sigma_y$: label standard deviation} & 1.5 & ~ & ~ & ~ & ~ & ~ & ~ \\  \hline
    \end{tabular}
}
\end{table*}

\bibliography{bieri_741}